\documentclass[a4paper,10pt,final]{llncs}  

\usepackage{boxedminipage}
\usepackage{graphicx}
\usepackage{algorithm}
\usepackage[noend]{algorithmic}
\usepackage{url}
\usepackage[colorlinks,linkcolor=blue,hyperindex]{hyperref}

\hypersetup{colorlinks,
   debug=false,
   linkcolor=blue,  
   citecolor=red,   
   urlcolor=blue,   
   bookmarksopen=true,
   pdftitle={Frame Interpretation and Validation in a Open Domain Dialogue System},
   pdfauthor={Artur Ventura, Nuno Diegues, David Martins de Matos},
   pdfsubject={Dialogue Frame Interpretation},
   pdfkeywords={Open domain dialogue, Frame interpretation, ECAs}
}

\begin{document}

\title{Frame Interpretation and Validation in a Open Domain Dialogue System}
\author{Artur Ventura, Nuno Diegues, David Martins de Matos}
\institute{L$^2$F -- Spoken Language Systems Laboratory\\
INESC ID Lisboa, Rua Alves Redol 9, 1000-029 Lisboa, Portugal\\
\url{{artur.ventura,nuno.diegues,david.matos}@l2f.inesc-id.pt}}
\maketitle

\pagestyle{plain}
\pagenumbering{arabic}

\begin{abstract}
Our goal in this paper is to establish a means for a dialogue platform to be able to cope with open domains considering the possible interaction between the embodied agent and humans. To this end we present an algorithm capable of processing natural language utterances and validate them against knowledge structures of an intelligent agent's mind. Our algorithm leverages dialogue techniques in order to solve ambiguities and acquire knowledge about unknown entities.
\end{abstract}

  %
  %

\section{Introduction}
\label{sec:intro}

Dialog systems and knowledge representations typically associated with agent systems have been merged in many situations due to their proximity in dealing with reasoning and human interaction. It has become a natural step to embody agents in robots deployed in the real world to either serve human requests or entertain them as companions. Our goal in this paper is to establish a means for a dialogue platform to be able to cope with open domains considering the possible interaction between the embodied agent and humans.

Our objective is to be able to interpret and validate the natural language utterances provided to the system against the agent's internal world model.

This document is organized as follows. In section~\ref{sec:work}, we present related work and context relevant to our work. In section~\ref{sec:open-domain}, we explain our approach in terms of linguistic and knowledge structure that will enable us to support situatedness. In
section~\ref{sec:interp}, we present the interpretation algorithms and, in section~\ref{sec:scenarios}, we showcase frame interpretation and validation in open domains in two scenarios. Finally, section~\ref{sec:final} presents our conclusions and directions for future developments.

\section{Related Work}
\label{sec:work}

As a consequence of the process of embodiment, there is a rising need of adequateness to the context the embodied agent is currently in and the ones it has experienced and related to which it has associated memories. Li et al.~\cite{Li:2009:CMM:1654595.1654626} evaluated the importance of situatedness in dialogue when the system is using embodied agents. In order to support situated dialogue, work on generation and resolution of referring expressions has been accomplished based on vision, in which the dialogue system depends on input from a vision subsystem to allow a reference resolver, along with spatial reasoning, to match linguistic references to world entities~\cite{Kelleher:2009:ACM:1596839.1596844}. Further experiments by Zender et al.~\cite{Zender:2009:SRG:1661445.1661703} used a bidirectional layer model for resolution and generation of referring expressions for entities that might not be in the current context and therefore must be accounted for as such when producing and interpreting dialogue in a human-robot interaction. Lison and Kruijff~\cite{Lison:2008:SCP:1567281.1567419} proposed a solution for dialogue systems to cope with open domains through priming speech recognition based on the concept of salience, from both linguistic and visual points of view. This concept was also a main focus target of Kelleher and Costello~\cite{Kelleher:2009:ACM:1596839.1596844}.

Systems where dialogue and agents come together have been referred to as conversational service robots, when they are meant to serve human requests, and conversational entertainment robots, when they focus on emotions display and human entertainment~\cite{nakano2005}. In Section~\ref{sec:interp}, we propose a means to abstract from such specification by showing that one can deal with open domains as long as all knowledge is linguistically annotated. In our approach we will refer to the agent's abilities as competencies. The competencies abstract the execution of specific actions. Since our work is closely related to the LIREC project's\footnote{\url{http://www.lirec.eu/}} architecture and objectives, we will consider competencies as part of the middle-layer of the LIREC architecture. This layer lies between the interface with the physical world and the deliberative mind and agent's memory. It is on this latter layer that we will focus our work. Similarly to this concept of competencies, Nakano et al.~\cite{nakano2005} suggested that their system's behaviour was based on modules, called experts, which would take charge of the interaction with the speaker, according to the domain inferred from the dialogue. Ultimately, these experts would carry on the interpretation of the utterances to an action that was physically performed, such as carrying on a request made by the speaker to search for an object. However, such an approach narrows the range of actions the system can perform, due to the strict connection of the modules to a physical entity such as an engine.

On the other hand, Bohus and Horvitz~\cite{Bohus:2009:DOW:1647314.1647323} proposed an open-world platform which attempts to allow a dialogue system to support multi-dynamic user interaction along with heavily situated context information acquired, mostly from vision features, to adapt the dialogue domain. We show that open domain dialogue adaptation can also rely on linguistic information rather than there approach, which focused mainly on visual information.

  %
  %

\section{Open Domain}
\label{sec:open-domain}

Open domains are sets of entities and relations containing several themes. If they interact with an external world, these domains can grow and change over time.

Typical frame-based dialogue systems normally operate over a set of entities and themes in very specific domains. These systems may be useful
for a small set of tasks with a well-defined number of entities (buying tickets, controlling a house, etc.) but are unable to deal with open domains.

Open Domain Dialog Systems (ODDS) use open domains and are, thus, capable of referring to a very large number of concepts. This also means that even with a small number of possible tasks, polysemy phenomena can be a problem. For instance, suppose that an ODDS has the capability of both finding spherical objects on a given space and buying items on online stores. Asking such a system to \textit{find a blue ball} can have multiple senses (e.g. finding a blue spherical objects or acquiring a Union musket ball from the American Civil War). In order to use frames in such a system, it is necessary to create an ontological model that merges linguistic information with world knowledge.

\subsection{Linguistic Information}
\label{sec:ling-info}

The linguistic information in the system's memory can be multi-lingual. In each language, word senses can have semantic relations with other
senses such as synonymous, hypernyms, among others. These relations make each language a linguistic ontology. For this we used WordNet~\cite{Miller:1995:WLD:219717.219748,fellbaum1998wordnet}.

Also, each language possesses a grammar description: each verb on the language has an associated set of structural (NP VP NP) and semantic
(Agent V Object) relations between itself and its arguments. We call this structure a Frame. Information for the Frames was obtained from~\cite{journals/lre/KipperKRP08}. Given that a verb can have multiple structural representations and senses, there must be an association between senses and frames which we call FrameSet. This association allows a semantic separation, for instance, between ``to find'' (acquiring) and ``to find'' (discovering). An example of these relations is presented in Figure~\ref{fig:rels}.

Each utterance given by the user is processed by a natural language processing chain. The result of this chain is a syntactic structure. Finally, this structure can also support associations between senses and its parts.

\subsection{Knowledge Organization}

In addition to linguistic information, the dialogue system contains information about the concepts that it can reason about. Information in the system's memory may contain concepts like physical objects, colors, locations, or geometry information. These entities will be matched against what the user said, in the process of obtaining a meaning for the sentence.

Furthermore, the system itself also describes itself as well as its set of competencies. Competencies correspond to interactions or abstractions of sensors and actuators in the agent's physical body (robot). Each of these competencies is described by its name, its actions, and its results. This way, it is not only possible to match abstract concepts to tangible actions, but it also becomes possible to speak about concepts for which the system does not have a formal definition. This can be seen in the following example: consider a user who asks for an object not described in the agent's memory. In this scenario, the system would not be able to ground the linguistic concepts to entities in the described world. It would require a definition which, if provided by the user, would be evaluated and matched against known concepts in the memory. After acquiring all the needed information, it would be capable of combining a set of competencies which would act cooperatively, based on each individual property on the new definitions , for that specific purpose.

\begin{figure}[ht]
\centering
\includegraphics[width=.62\columnwidth]{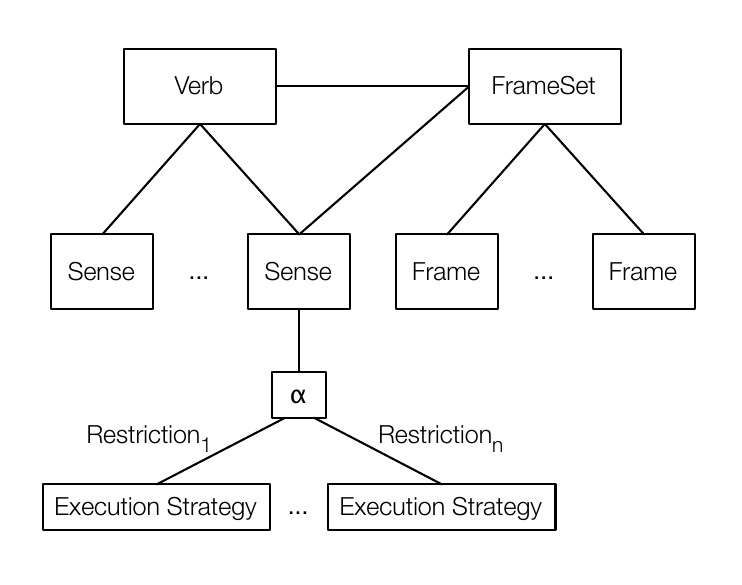}
\caption{Relations between Verb, Sense, FrameSet, Frame and $\alpha$-structure.}
\label{fig:rels}
\end{figure}

Verb senses that are meaningful for the system (i.e. it is possible to create a plan for them) are also associated with execution strategies with restrictions. We call this association an $\alpha$-structure as shown in figure~\ref{fig:rels}.

$\alpha$-structures allow the separation of equal senses based on restrictions in different execution strategies: \textit{find something} may represent a sense of searching for something, but the act of finding an physical object in a space or finding a person in a building are essentially different tasks requiring different plans.

Execution strategies are abstract plans that can serve different purposes, depending on their arguments. When the interpretation algorithm is executed, the $\alpha$-structure will be associated with the utterance's verb. Restrictions allow choosing an execution strategy, based on the verb arguments and on the context. As an example, consider searching for either a ball or a rubber duck: it may be essentially the same task, but the sensors and actuators required may be different. The instantiation of this execution strategy will be an executable plan.

\subsection{Language and Knowledge Bridge}

To enable the use of a frame system in an open domain it was necessary to develop an ontology memory model that merges the agent's world
knowledge with linguistic information.

In this domain representation, all ontology components are represented by OL nodes. These nodes contain two layers:

\begin{itemize}
  \item The O layer contains a non-descriptive representation of a component. This layer maintains an entity typification and its relations
with other entities.
  \item The L layer contains a list of senses. Each sense is associated with one language, as described in Section~\ref{sec:ling-info}.
\end{itemize}

\begin{figure}[ht]
\centering
\includegraphics[width=.9\columnwidth]{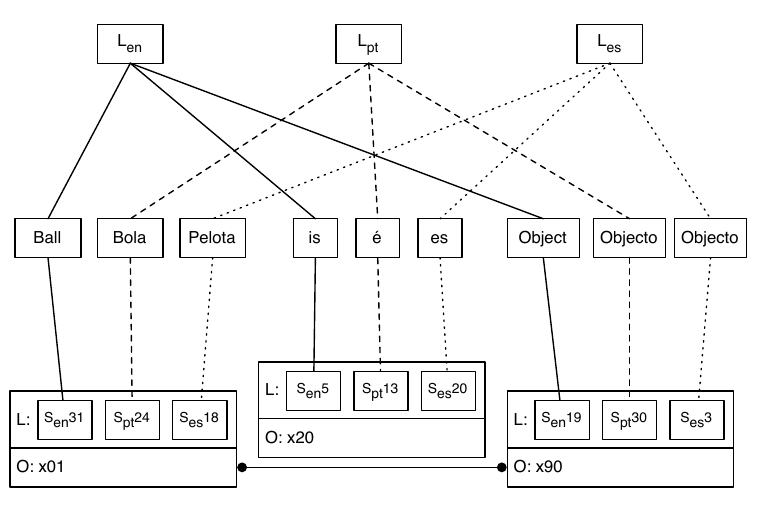}
\caption{Ontological and linguistic representation of the relationship between ``Ball'' and ``Object''.}
\label{fig:bridge}
\end{figure}

The O-ontology represents all the concepts that the dialogue system knows about and is able to handle. In this layer, the concept of a ball would be connected to the notion of physical object, as shown in figure~\ref{fig:bridge}. This object may have other properties, such as color or size. The L layer would allow intersections between each concept and their description in a given language. This way, different concepts could share equal words without the danger of ambiguity.

  %
  %

\section{Interpretation and Validation of Frames}
\label{sec:interp}

When a new utterance is detected by the system, this is initially processed by the natural language processing chain. The resulting tree, containing the verb and its arguments, is passed along with the language in which it was processed to \textsc{Interpret} (algorithm~\ref{fig:algo-interp}). This algorithm matches what was said to a meaningful structure in the system memory. In this algorithm, a list of FrameSets is obtained from the sentence's verb. For each member of this list, \textsc{Sound} determines if the sentence structure matches some of the Frames in the FrameSet. If it does, all the possible meanings obtained by the combination of word senses are going to be generated.

\begin{algorithm}
\caption{\textsc{Interpret} algorithm.}
\label{fig:algo-interp}
\begin{algorithmic}[1]
\STATE INTERPRET($t$,$l$):
\STATE $lFrameSet \leftarrow$ FRAMESETS(VERB($t$), $l$)
\STATE $r \leftarrow []$
\FORALL{$fs \in lFrameSet$}
\IF{SOUND($fs$, $t$)}
\STATE $lMeaning \leftarrow$ MEANINGS($fs$, $t$, $l$)
\STATE $f \leftarrow$ FRAME($fs$, $t$)
\FORALL{$m \in lMeaning$}
\FORALL{$es \in$ STRATEGIES($m$, $f$)}
\IF{VALID($es$, $m$)}
\STATE PUSH($r$,INSTANTIATE($es$, $m$))
\ENDIF
\ENDFOR
\ENDFOR
\ENDIF
\ENDFOR
\RETURN $r$
\end{algorithmic}
\end{algorithm}

\subsection{Generating Meaning Combinations}

\textsc{Combinations} (algorithm~\ref{fig:algo-combin}) creates a list of tree copies with all possible combinations of senses for
the verb arguments. An SK is going to query the memory for all the senses of each argument. If an argument is a compound word (e.g. \textit{the blue ball}) and it is not represented in memory, a structure is created in memory containing the combination of all the words. The latter would mean that this structure for our example would be associated with the concepts \textit{blue} and \textit{ball}.

\begin{algorithm}
\caption{\textsc{Combinations} algorithm.}
\label{fig:algo-combin}
\begin{algorithmic}[1]
\STATE COMBINATIONS(t,l):
\STATE $r \leftarrow [t]$
\FORALL{$arg \in$ ARGS($t$)}
\STATE $temp \leftarrow []$
\FORALL{$ti \in r$}
\STATE $known \leftarrow$ ASK($arg$, $l$)
\IF{LENGTH($known$) $= 0$}
\STATE $known \leftarrow$ INQUIRY($arg$, $l$)
\ENDIF
\FORALL{$s \in known$}
\STATE $ti \leftarrow $ COPY($ti$)
\STATE SET-SENSE($arg$, $ti$, $s$)
\STATE PUSH($temp$, $ti$)
\ENDFOR
\ENDFOR
\STATE $r \leftarrow temp$
\ENDFOR
\RETURN $r$
\end{algorithmic}
\end{algorithm}

If no sense is found for an argument, meaning that this concept is not represented in memory or a mapping between this language and this concept does not exist, an INQUIRY is called for the argument. This action will suspend the current computation and probe for a sense: this can be achieved by querying the user for the sense, or executing aknowledge augmentation algorithm over the memory. Once a valid sense has been obtained for the argument, the computation will resume.

\textsc{Combinations} returns a list with trees annotated with senses. This list must then be combined with all the $\alpha$-structures provided by the current FrameSet senses. This is done by \textsc{Meanings} (section~\ref{sec:matching}). The final list contains all possible meanings that the memory can provide for that sentence.

\subsection{Matching Meaning with Valid Actions}
\label{sec:matching}

After generating frame candidates in the previous steps, each of the elements in the list returned by \textsc{Meanings} (algorithm~\ref{fig:algo-meaning}) will be validated. For every execution strategy in an element, associated restrictions will be matched against the argument senses of that element. If they can be matched, the execution plan is instantiated and the result is collected. If not, it is discarded.

\begin{algorithm}
\caption{\textsc{Meanings} algorithm.}
\label{fig:algo-meaning}
\begin{algorithmic}[1]
\STATE MEANINGS($fs$,$t$,$l$):
\STATE $r \leftarrow []$
\FORALL{$s \in$ SENSES($fs$)}
\STATE $\alpha\leftarrow$ FIND-$\alpha$($s$)
\FORALL{$t_i \in$ COMBINATIONS($t$, $l$)}
\STATE SET-SENSE(VERB($t_i$), $t_i$ , $\alpha$)
\STATE PUSH($r$, $t_i$)
\ENDFOR
\ENDFOR
\RETURN $r$
\end{algorithmic}
\end{algorithm}

If the result list of \textsc{Interpret} is unitary, then there is only one possible interpretation. If contains more than one element, then we have an ambiguity. The system can choose one of them, or ask the user what to do. If the list is empty, the system understood all the concepts, but no action could be taken.

  %
  %

\section{Scenarios}
\label{sec:scenarios}

We consider two scenarios to illustrate our method and how it handles interpretation and validation problems. The first scenario considers a situation in which two possible execution strategies exist but only one is valid. The second scenario considers an ambiguous request.

\subsection{Scenario 1: Jacob}

Suppose that our entity is called \textit{Jacob} and it knows concepts like ball and the color blue. These are connected with the WordNet concepts \textit{ball} (noun) and \textit{blue} (adjective).

Verb \textit{find} is associated with a FrameSet populated with information from the VerbNet. One of the Frames in this FrameSet contains the structural description NP V NP and semantic Agent V Theme. This frame is connected to senses from that verb. One of those, the sense of discovery (\verb|find%2:39:02::|) is associated with an $\alpha$-structure that contains two execution strategies. The first of these requires the Theme to be a person, to be located in a physical location, and the Agent to have a physical actuator that allows mobility. The second execution strategy also requires the Agent to have a physical actuator that allows mobility, the Theme to be a physical object and the Agent must have the capability to detect Theme objects.

If the user says \textit{Jacob find the blue ball}, the initial syntactic chain would return a structure associating NP to \textit{Jacob}, V to \textit{find} and \textit{the blue ball} to NP (containing an adjective \textit{blue} and a noun \textit{ball}). This structure is then passed to the INTERPRET algorithm.

When generating the possible combinations, the first argument will match with the entity representing the system. The other argument, \textit{the blue ball}, will have 96 possible senses, since WordNet presents 8 for \textit{blue} and 12 for \textit{ball}. However, we only know one of these, the physical ball combined with the color blue. So, the only combination returned from \textsc{Combinations} is:

\centerline{\includegraphics[width=.5\columnwidth]{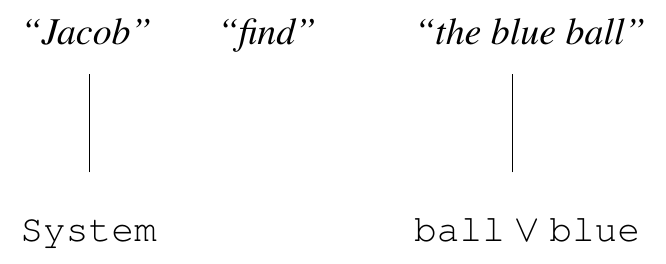}}

\textsc{Meanings} will also return a list containing only one meaning because there is only one $\alpha$-structure in the senses for that FrameSet:

\centerline{\includegraphics[width=.5\columnwidth]{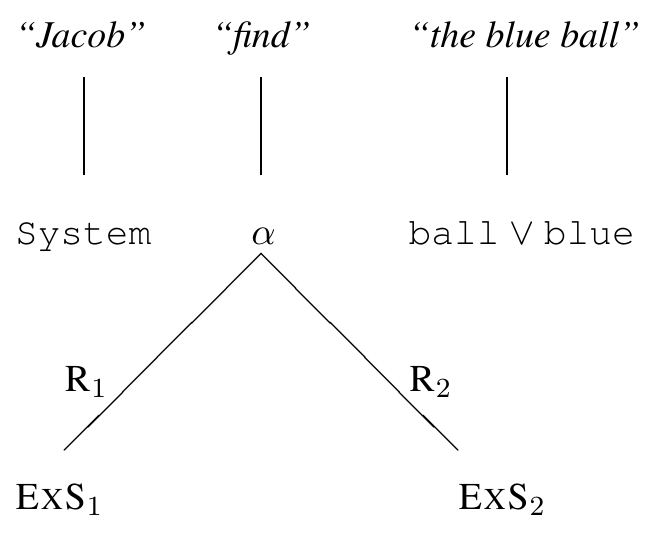}}

When the execution strategies are verified for restrictions, the only valid meaning has the following associations:

\centerline{\includegraphics[width=.62\columnwidth]{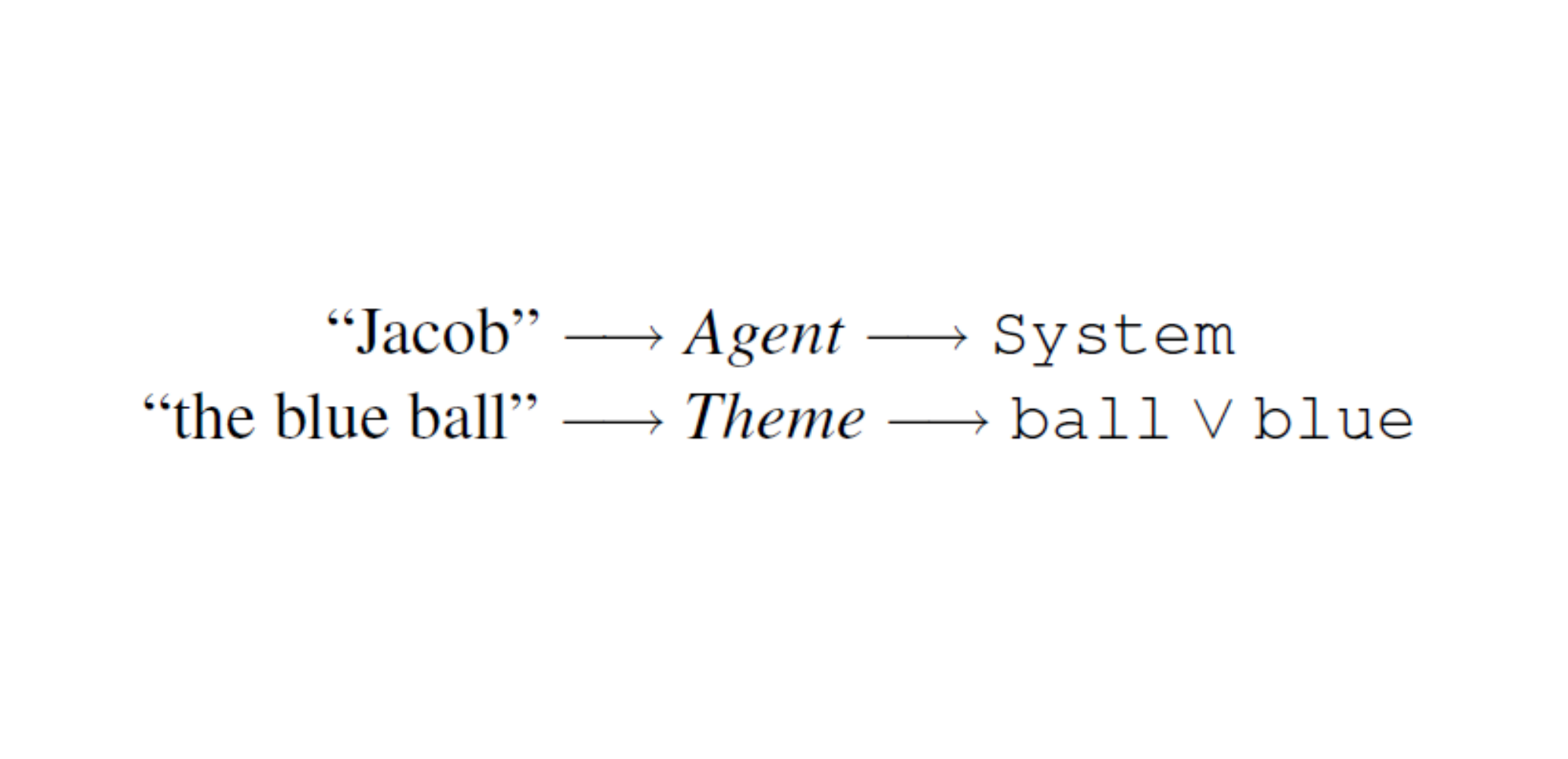}}

The $\alpha$-structure has two execution associated strategies. The restrictions for the first will fail because \textit{the blue ball} is not a person, but the second may be valid if the system has a detector of colored balls. Supposing it does, this causes \textsc{Interpret} to return a plan for that action.

\subsection{Scenario 2: Motors}

Now let us suppose that the system has described finegrained competencies that allow it to control specific internal motors and has the concept of that motor described in its memory. Furthermore, suppose the system has competencies capable of starting and stopping external motors.

In this scenario, we ask to the system \textit{Jacob start motor nine}. \textsc{Meanings} will return two possible meanings, one for each of the motors. The verb sense for these different tasks is the same (initiating something), so only one of the $\alpha$-structures will be found.

Two execution strategies will be associated with $\alpha$-structure, but both will be valid since for each one there is a valid meaning. Accordingly, \textsc{Interpret} will return two plans, one for each of motor. In this case the system proceeds as described in section~\ref{sec:matching}.

  %
  %

\section{Discussion and Final Remarks}
\label{sec:final}

We presented a set of algorithms capable of processing natural language utterances and validate them against knowledge structures of an intelligent agent's mind. Our algorithm provides a means for activating plans to prompt the user through dialogue when faced with ambiguous situations or situations with insufficent information.

Even though our algorithms are able to fulfil our objectives, we are aware that they requires a high capability for describing language at multiple levels (eg. morphosyntatic, syntatic, ontological). Language resources for accomplishing this goal may not easily available for all languages one could think of using in an agent, thus increasing the cost, or even preventing the use, of our solution. Nevertheless, we believe that in time these factors will cease to be a problem as more resources are made available by the community.

We plan to support automatic linguistic annotation of ontological knowledge available in the agent's mind. This feature is useful since we envision situations in which we are able to enlarge the agent's knowledge through pluggable ontologies. We believe that linguistic information for this new knowledge can be easily acquired through dialogue interactions.

\section*{Acknowledgements}

This work was partially supported by European Project LIREC (FP7/2007-2013 under grant agreement \#215554).

\bibliographystyle{plain}
\bibliography{document}

  %
  %

\end{document}